\documentclass[letterpaper, 10 pt, conference]{ieeeconf}  

\IEEEoverridecommandlockouts

\usepackage[utf8]{inputenc}
\usepackage{amsmath, scalerel} 
\usepackage{amssymb}
\usepackage{amsopn}
\usepackage{graphicx}
\usepackage[style=base]{caption}
\usepackage{color}
\usepackage{booktabs}
\usepackage{todonotes}
\usepackage{lipsum}
\usepackage{import}
\usepackage{footnote}
\usepackage{multirow}
\usepackage{hhline}
\usepackage{subcaption}
\usepackage[pdfborder={0 0 0}, bookmarks=false, final]{hyperref}

\usepackage{floatrow}

\usepackage{nicefrac}

\usepackage[abbreviations,binary-units]{siunitx}
\DeclareSIUnit{\mAh}{mAh}
\DeclareSIUnit{\Wh}{Wh}

\usepackage[gernotes]{optional}

\usepackage{comment}
\excludecomment{confidential}

\usepackage{pgfplots}
\pgfplotsset{compat=newest}
\usepgfplotslibrary{patchplots}
\usepackage{grffile}
\usepackage{ textcomp }

\usepackage{pgfplots}
\usepackage{pgf}
\pgfplotsset{compat=newest} 
\pgfplotsset{plot coordinates/math parser=false} 
\newlength\figureheight 
\newlength\figurewidth 

\title{\LARGE \bf
A Perceived Environment Design using a Multi-Modal Variational Autoencoder for learning Active-Sensing
}

\author{Timo Korthals$^{\dagger *}$, Malte Schilling$^{\dagger}$, Jürgen Leitner$^{\text{\textdaggerdbl}}$
\thanks{
	$^{\dagger}$Bielefeld University,
	Cluster of Excellence Cognitive Interaction Technologies,
	Inspiration 1, 33619 Bielefeld, Germany\newline
	\indent {$^\text{\textdaggerdbl}$}JL is with the Australian Centre for Robotic Vision,
	Queensland University of Technology,
	Brisbane, Australia\newline
	\indent{$^{*}${\tt\small tkorthals@cit-ec.uni-bielefeld.de}}\newline
	\indent{}This research was supported by the Federal Ministry of Education and Research (57388272), and by 'CITEC' (EXC 277) at Bielefeld University which is funded by the German Research Foundation (DFG). The responsibility for the content of this publication lies with the author.
	}
}

\makeatletter
\newcommand{\myspecialcell}[1]{\ifmeasuring@#1\else\omit$\displaystyle#1$\ignorespaces\fi}
\makeatother

\usepackage{tabularx}

\usepackage{caption}
\begin{document}

\maketitle
\thispagestyle{empty}
\pagestyle{empty}

\section{INTRODUCTION}
\label{sec:intro}
\textit{Active sensing} (AS) is one of the most fundamental problems and challenges in mobile robotics which seeks to maximize the efficiency of an estimation task by actively controlling the sensing parameters \cite{Mihaylova2002}.
AS can be divided into two sub-tasks:
the identification of a \textit{point of interest} (PoI) to achieve (e.g. object to sense, estimation of vantage points, or selection of sensing modalities) and the ability of a robot to navigate through the environment to reach a certain goal location.
In teams of heterogeneous robots that employ different sensory modalities, AS is of particular interest
as it can be used to resolve observation ambiguities.

Friston \cite{Friston2010} states that minimizing free energy is equivalent to maximizing model evidence, which is equivalent to minimizing the complexity of accurate explanations for observed outcomes.
Following this principle, if one could directly obtain an estimation of free energy through the current observation, a controller for sensing parameters can be learned that minimizes free energy.
This approach would enable intrinsically motivated training, comparable to curiosity driven learning as proposed by Pathak et al. \cite{DBLP:journals/corr/PathakAED17}, which facilitates the autonomous gathering of information about the environment.
Moreover, it would enable an epistemic (ambiguity resolving) goal-directed behavior, as it would only take the effort to approach a particular sensor configuration if, and only if, it helps to resolve a observation ambiguity.

This issue was addressed by Korthals et al. \cite{korthals2019icra}, who proposed a novel approach to AS, which intrinsically reduces ambiguities of observations through epistemic actions.
They facilitated the correlation between maximization of the trainable evidence lower bound (ELBO), through a \textit{multi-modal variational autoencoder} (M\textsuperscript{2}VAE), with the minimization of variational free energy, resulting in active-sensing with epistemic behaviors.
Every sensing agent was trained by applying a \textit{deep reinforcement learning} (DRL) approach utilizing separate deep Q-learning \cite{Mnih2015}, such that every agent learns the interplay between state, action, and reward.

This contribution focuses on a detailed and extended description of the architecture used in \cite{korthals2019icra}.
We comprise the interplay between the M\textsuperscript{2}VAE and an \textit{environment} to a \emph{perceived environment}, on which an agent can act, in Section \ref{sec:approach}.
Then we conclude with a comparison to curiosity-driven learning in Section \ref{sec:results}. 

\section{APPROACH}
\label{sec:approach}
\begin{figure*}[!ht]
	\def\svgwidth{\textwidth}
	\footnotesize
\begingroup
  \makeatletter
  \providecommand\color[2][]{
    \errmessage{(Inkscape) Color is used for the text in Inkscape, but the package 'color.sty' is not loaded}
    \renewcommand\color[2][]{}
  }
  \providecommand\transparent[1]{
    \errmessage{(Inkscape) Transparency is used (non-zero) for the text in Inkscape, but the package 'transparent.sty' is not loaded}
    \renewcommand\transparent[1]{}
  }
  \providecommand\rotatebox[2]{#2}
  \newcommand*\fsize{\dimexpr\f@size pt\relax}
  \newcommand*\lineheight[1]{\fontsize{\fsize}{#1\fsize}\selectfont}
  \ifx\svgwidth\undefined
    \setlength{\unitlength}{481.88976378bp}
    \ifx\svgscale\undefined
      \relax
    \else
      \setlength{\unitlength}{\unitlength * \real{\svgscale}}
    \fi
  \else
    \setlength{\unitlength}{\svgwidth}
  \fi
  \global\let\svgwidth\undefined
  \global\let\svgscale\undefined
  \makeatother
  \begin{picture}(1,0.35305295)
    \lineheight{1}
    \setlength\tabcolsep{0pt}
    \put(0,0){\includegraphics[width=\unitlength,page=1]{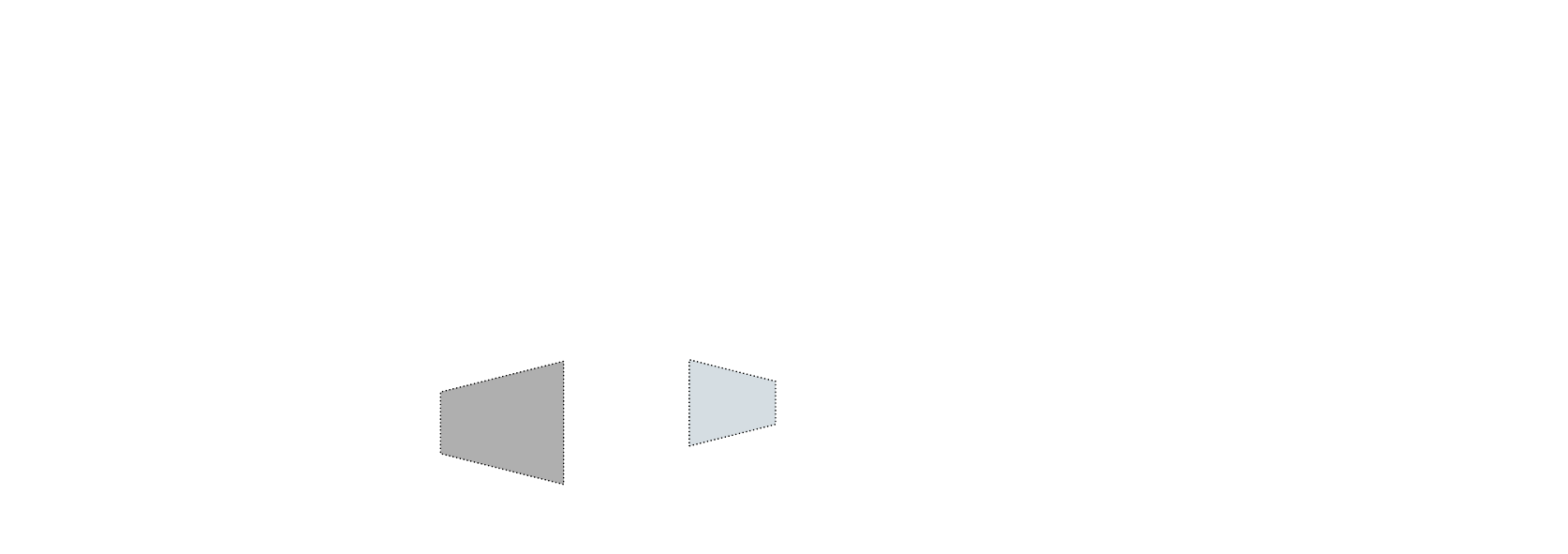}}
    \put(0.5661236,0.10406314){\color[rgb]{0,0,0}\makebox(0,0)[lt]{\begin{minipage}{0.28992501\unitlength}\raggedright $z_t$\end{minipage}}}
    \put(0.39553427,0.1217703){\color[rgb]{0,0,0}\makebox(0,0)[lt]{\begin{minipage}{0.28992501\unitlength}\raggedright $O_{t}$\end{minipage}}}
    \put(0.23373819,0.02211899){\color[rgb]{0,0,0}\makebox(0,0)[lt]{\begin{minipage}{0.31793971\unitlength}\raggedright $f\left(S_{t+1}|S_{t},A_{t},z_{t+1}\right)$\end{minipage}}}
    \put(0.24259419,0.08911569){\color[rgb]{0,0,0}\makebox(0,0)[lt]{\begin{minipage}{0.28992501\unitlength}\raggedright $z_{t+1}$\end{minipage}}}
    \put(0.24138017,0.17110806){\color[rgb]{0,0,0}\makebox(0,0)[lt]{\begin{minipage}{0.28992501\unitlength}\raggedright $r\left(z_{t},z_{t+1}\right)$\end{minipage}}}
    \put(0.12611875,0.19979983){\color[rgb]{0,0,0}\makebox(0,0)[lt]{\begin{minipage}{0.28992501\unitlength}\raggedright \textbf{Perceived Environment}\end{minipage}}}
    \put(0.60142929,0.18978454){\color[rgb]{0,0,0}\makebox(0,0)[lt]{\begin{minipage}{0.48447157\unitlength}\raggedright \textit{Environment} with $K=3$ robots and $N=4$ PoIs\end{minipage}}}
    \put(0,0){\includegraphics[width=\unitlength,page=2]{architecture_rl.pdf}}
    \put(0.48267271,0.06995049){\color[rgb]{0,0,0}\makebox(0,0)[lt]{\begin{minipage}{0.28992501\unitlength}\raggedright $O_{t+1}$\end{minipage}}}
    \put(0,0){\includegraphics[width=\unitlength,page=3]{architecture_rl.pdf}}
    \put(0.56103602,0.34779617){\color[rgb]{0,0,0}\makebox(0,0)[lt]{\begin{minipage}{0.28992501\unitlength}\raggedright \textbf{Meta-Agent}\end{minipage}}}
    \put(0,0){\includegraphics[width=\unitlength,page=4]{architecture_rl.pdf}}
    \put(0.05459942,0.12488305){\color[rgb]{0,0,0}\makebox(0,0)[lt]{\begin{minipage}{0.28992501\unitlength}\raggedright $R_{t+1}$\end{minipage}}}
    \put(0.05459942,0.06723747){\color[rgb]{0,0,0}\makebox(0,0)[lt]{\begin{minipage}{0.28992501\unitlength}\raggedright $S_{t+1}$\end{minipage}}}
    \put(0.18401118,0.27942064){\color[rgb]{0,0,0}\makebox(0,0)[lt]{\begin{minipage}{0.28992501\unitlength}\raggedright $R_{t}$\end{minipage}}}
    \put(0.18401118,0.326061){\color[rgb]{0,0,0}\makebox(0,0)[lt]{\begin{minipage}{0.28992501\unitlength}\raggedright $S_{t}$\end{minipage}}}
    \put(0.70165824,0.3142963){\color[rgb]{0,0,0}\makebox(0,0)[lt]{\begin{minipage}{0.28992501\unitlength}\raggedright $A_{t}$\end{minipage}}}
    \put(0.45258806,0.10080993){\color[rgb]{0,0,0}\makebox(0,0)[lt]{\begin{minipage}{0.28992501\unitlength}\raggedright Dec.\end{minipage}}}
    \put(0.30935583,0.08765123){\color[rgb]{0,0,0}\makebox(0,0)[lt]{\begin{minipage}{0.28992501\unitlength}\raggedright Enc.\end{minipage}}}
    \put(0,0){\includegraphics[width=\unitlength,page=5]{architecture_rl.pdf}}
    \put(0.47074615,0.29557866){\color[rgb]{0,0,0}\makebox(0,0)[lt]{\lineheight{1.25}\smash{\begin{tabular}[t]{l}Shared\\Network\end{tabular}}}}
    \put(0,0){\includegraphics[width=\unitlength,page=6]{architecture_rl.pdf}}
    \put(0.46575931,0.22432841){\color[rgb]{0,0,0}\makebox(0,0)[lt]{\lineheight{1.25}\smash{\begin{tabular}[t]{l}multi-headed DQN\end{tabular}}}}
    \put(0.5500034,0.24542452){\color[rgb]{0,0,0}\makebox(0,0)[lt]{\lineheight{1.25}\smash{\begin{tabular}[t]{l}$M=2$ heads\end{tabular}}}}
    \put(0,0){\includegraphics[width=\unitlength,page=7]{architecture_rl.pdf}}
    \put(0.72899483,0.06750663){\color[rgb]{0,0,0}\makebox(0,0)[lt]{\begin{minipage}{0.28992501\unitlength}\raggedright a\end{minipage}}}
    \put(0.78550653,0.10118179){\color[rgb]{0,0,0}\makebox(0,0)[lt]{\begin{minipage}{0.28992501\unitlength}\raggedright b\end{minipage}}}
    \put(0.68881489,0.11556424){\color[rgb]{0,0,0}\makebox(0,0)[lt]{\begin{minipage}{0.28992501\unitlength}\raggedright 1\end{minipage}}}
    \put(0.70092059,0.04403034){\color[rgb]{0,0,0}\makebox(0,0)[lt]{\begin{minipage}{0.28992501\unitlength}\raggedright 2\end{minipage}}}
    \put(0.82031912,0.12420922){\color[rgb]{0,0,0}\makebox(0,0)[lt]{\begin{minipage}{0.28992501\unitlength}\raggedright 3\end{minipage}}}
    \put(0,0){\includegraphics[width=\unitlength,page=8]{architecture_rl.pdf}}
    \put(0.768104,0.04879767){\color[rgb]{0,0,0}\makebox(0,0)[lt]{\begin{minipage}{0.28992501\unitlength}\raggedright c\end{minipage}}}
    \put(0,0){\includegraphics[width=\unitlength,page=9]{architecture_rl.pdf}}
    \put(0.8294713,0.05185856){\color[rgb]{0,0,0}\makebox(0,0)[lt]{\begin{minipage}{0.28992501\unitlength}\raggedright 4\end{minipage}}}
  \end{picture}
\endgroup

	\vspace{-1em}
	\caption{Overview of the meta-agent--environment interaction. The \textit{perceived environment} consists of the simulator that executes actions and returns sensor readings, and a multi-modal variational autoencoder structure as proposed in \cite{Korthals2019FUSION} for calculating the reward and observation embedding. The meta-agent is requested $K$ times for each robot, with the corresponding modality dependent head network to build the joint action $A\in\mathcal{A}^{K}$.\vspace{-2em}}
	\label{fig:VAE}
\end{figure*}
The \emph{perceived environment} (c.f. Fig. \ref{fig:VAE}) is an environment with observation post-processing that consists of four parts: the original \emph{environment} (simulator or real-world), the state transition function $f$, the deep generative model M\textsuperscript{2}VAE, and the reward function $r$.
The original \emph{environment} manages $K$ robots equipped with one of $M$ sensing modalities.
It executes control action $A_t \in \mathcal{A}$ for one robot $k$, and generates its sensor observations $O_{t+1}$.
An action $A$ for some chosen robot $k$ lets it drive and observe PoI $n$ in the environment.
The M\textsuperscript{2}VAE was pre-trained as described in \cite{Korthals2019FUSION} on $M$ modalities with Gaussian parameter layers in the bottleneck that produce an embedding $z=\left(\mu,\sigma\right)$ (Notice that $z$ is a tuple of statistical parameters).
The encoder networks of the M\textsuperscript{2}VAE are used to encode a current observation $O_{t+1}$, while its decoder networks are used to decode a former embedding $z_t$ to $O_{t}$ that can be fused via re-encoded with the current observations.
The state transition function $f(S_{t+1} \mid S_t, A_t, z_{t+1})$ produces the new state $S_{t+1}$ based on the taken action $A_t$, the former state $S_t$, and the new embedding $z_{t+1}$.
The reward function $r$ calculates, based on the shift between the observations' embeddings $z_t$ and $z_{t+1}$ the reward $R_{t+1}$.
Unlike the reward function defined by Korthals et al. \cite{korthals2019icra}, we based a new definition of reward on the \textit{epistemic value} defined by Friston et al. \cite{friston2017} which is the mutual information between hidden states: $r\left(z_t , z_{t+1}\right) \propto D_{\text{KL}}\left(z_t || z_{t+1}\right)$.
The \textit{epistemic value} reports the reduction in uncertainty about hidden states afforded by new observations by calculating the Kullback--Leibler divergence (KL) $D_{\text{KL}}$.
Because the KL or information gain cannot be less than zero, it disappears when the former embeddings are not informed by new observations.

The \textit{agent} (c.f. Fig. \ref{fig:VAE}) comprises a policy network that maps a state to action.
While a single \emph{deep Q-network} (DQN) was used as a meta-agent in \cite{korthals2019icra} to control multiple robots, we extended the meta-agent to an $M$-headed DQN that controls each robot differently depending on its sensor modality (depicted in Fig. \ref{fig:VAE}). 

Next, we define the internal state of the \textit{perceived environment}: $S=\left(Z,V,T_1, \ldots, T_K\right)$.
The environmental -- robot independent -- state of the world hold the M\textsuperscript{2}VAE's embeddings $Z= \left(\mu_1, \ldots, \mu_N, \sigma_{1}, \ldots, \sigma_{N}\right)$, and visits $V=\left(v_1, \ldots, v_N\right)$ for every PoI $n$.
A visit $v_n\in\left\lbrace0,1\right\rbrace^M$ indicates which modalities have already observed this particular spot.
PoIs can be stored by an environmental representation like a grid-cell or topological map where every cell -- or node respectively -- $n$ holds the information $\left(\mu_n, \sigma_{n}, v_n\right)$.
A robot dependent known pose $T_k$ is given by the simulator or any localization system.

The state representation $S_{t+1}$ passed to the policy network is constructed for every robot $k$ by $f$ as follows:
We decided the policy network had a fixed input size and, therefore, focus only on $I$ PoIs in the vicinity of a robot.
However, this approach is just quasi-myopic since only PoIs that have not been perceived by the same modality, indicated by $V$, were respected.
Therefore, the state's respected surroundings can grow greedily to an arbitrary size.
Thus, the robot specific state for the policy network comprises the world's state of $I$ out of $N$ PoI-embeddings and its path distances $D$ for the requested robot $k$ to every PoI: 
$\mathcal{S}_{t+1}\mid k=\left( \mu_1, \ldots, \mu_I, D_{1}, \ldots, D_{I} \right)$.

For the agent's network output space $\mathcal{A}$, we assumed that each PoI could be observed by taking one action $a_i$, or the episode could be terminated before observing all PoIs by the selection of \textit{no-operation} (NOP): $\mathcal{A}=\left( a_1, \ldots, a_I, \text{NOP} \right)$.
The agent's policy $\pi_m$ was calculated based on the shared network and the modality dependent head, which was always chosen with respect to the currently controlled robot.
PoIs were marked as visited if the policy samples NOP and the task was done if no more PoIs could be visited anymore.

We calculate an average reward over all agents as the team reward $\bar{r}=\sum_{K}r_{k}$ for all robot's to encourage cooperative behavior.
Thus, every agent followed the policy, which maximized the future expected reward for the team.

\section{Comparison \& Future Work}
\label{sec:results}
Inspired by active inference, our depicted architecture facilitates an unsupervised approach to learning active sensing in a heterogeneous, distributed sensor network.
We furthermore hypothesize that training a variational autoencoder to facilitate its latent representation and exploiting its statistics as a reward has significant advantages over the recently published curiosity-driven approach by Pathak et al. \cite{DBLP:journals/corr/PathakAED17}.
First, their proposed intrinsic curiosity module (ICM) only produces rewards during training and not after convergence (what is described as boredom).
Second, the ICM can only learn from observations which can be manipulated by actions.
While this is a beneficial feature to initially ease the training process, it prevents the learning of processes that can only be observed but not controlled or only controlled indirectly.
Therefore, first, the ICM's embeddings can actually hardly be used as an input to a policy network, and second, an extrinsic reward is always necessary to enforce richer behaviors.

This hypothesis encourages future work that compares our \textit{perceived environment} with the ICM.
Furthermore, we will merge our proposed two-step approach to an end-to-end learning architecture.

\bibliographystyle{IEEEtran}
\bibliography{root}

\end{document}